\documentclass[letterpaper, 10 pt, conference]{ieeeconf}  

\IEEEoverridecommandlockouts  
\usepackage{cite}
\usepackage{amsmath,amssymb,amsfonts}
\usepackage{algorithmic}
\usepackage{graphicx}
\usepackage{textcomp}
\usepackage{xcolor}
\usepackage{booktabs}
\usepackage{multirow}
\usepackage{caption}
\usepackage{subcaption}
\newcommand{\bhline}[1]{\noalign{\hrule height #1}}

\usepackage{url}
\usepackage{color}
\usepackage{bm}

\title{Enlarged Large Margin Loss for Imbalanced Classification}

\author{Sota Kato$^{1}$ and Kazuhiro Hotta$^{2}$
\thanks{$^{1}$Sota Kato is with Department of Electrical and Electronic Engineering, Meijo University, Nagoya, Japan
        {\tt\small 150442030@ccalumni.meijo-u.ac.jp}}%
\thanks{$^{2}$Kazuhiro Hotta is with Department of Electrical and Electronic Engineering, Meijo University, Nagoya, Japan
        {\tt\small kazuhotta@meijo-u.ac.jp}}%
}

\begin{document}

\maketitle
\thispagestyle{empty}
\pagestyle{empty}

\begin{abstract}
We propose a novel loss function for imbalanced classification. 
LDAM loss, which minimizes a margin-based generalization bound, is widely utilized for class-imbalanced image classification.
Although, by using LDAM loss, it is possible to obtain large margins for the minority classes and small margins for the majority classes, the relevance to a large margin, which is included in the original softmax cross entropy loss, is not be clarified yet.
In this study, we reconvert the formula of LDAM loss using the concept of the large margin softmax cross entropy loss based on the softplus function and confirm that LDAM loss includes a wider large margin than softmax cross entropy loss.
Furthermore, we propose a novel Enlarged Large Margin (ELM) loss, which can further widen the large margin of LDAM loss.
ELM loss utilizes the large margin for the maximum logit of the incorrect class in addition to the basic margin used in LDAM loss.  
Through experiments conducted on imbalanced CIFAR datasets and large-scale datasets with long-tailed distribution, we confirmed that classification accuracy was much improved compared with LDAM loss and conventional losses for imbalanced classification.

\end{abstract}

\section{INTRODUCTION}
In recent years, convolutional neural networks (CNNs) are widely applied to various image classification tasks with great success \cite{b13,b14}.
However, there is a problem that are great differences in the number of samples in each class when CNNs are applied to real-world datasets \cite{b1,b2}, and typically we call this a class-imbalanced problem.
In class-imbalanced image classification, images belonging to a certain class account for the majority of datasets, and those belonging to the remaining classes are the minority. 
Consequently, the accuracy of minority classes will be quite low because a classifier is trained on datasets with an imbalanced sample ratio between classes.
To improve the classification performance on imbalanced datasets, it is important to improve the accuracy of minority classes.

The most common strategy for learning imbalanced data is the re-weighting strategy \cite{b3,b4,b5,b6,b7,b8,b9,b10,b11,b12,b21}.
While various approaches for re-weighting strategy have been proposed in recent years, one of the most effective approaches is a margin-based loss function like Label-Distribution-Aware Margin (LDAM) loss \cite{b4}.
It can regularize the minority classes more strongly than the majority classes, and gain per-class and uniform-label test error bounds based on the theory in \cite{b24}, by accounting for the minimum margin per class. 
Subsequently, by minimizing the obtained bounds, we can improve the accuracy of minority classes without sacrificing the ability of the model to fit the frequent classes.
On the other hand, in a recent study, it has been revealed that the softmax cross entropy loss includes a large-margin effect which is dependent on the distribution of logits \cite{b15}.
Although LDAM loss is a loss function based on softmax cross entropy loss, the relevance between LDAM loss and Large-Margin Softmax Cross Entropy (LMSCE) loss \cite{b15} is not clarified yet.

Therefore, in this study, we reconvert a formula for LDAM loss in terms of a large-margin perspective.
Then, we confirm that LDAM loss includes wider and class-specific values of large margins than LMSCE loss.
In imbalanced learning, we consider this to be the reason that LDAM loss is a better performance than other losses based on softmax cross entropy loss.
Furthermore, inspired by the above analysis, we propose a novel Enlarged Large Margin (ELM) loss with even wider margins than LDAM loss.
A stronger regularization than LDAM loss can be achieved by further increasing the margins between the logit of the target class and the logit of the class with the largest logit among the incorrect classes.
Consequently, we can achieve better-balanced learning than LDAM loss even if the dataset has a class-imbalanced distribution.

Through experiments, we evaluated our loss function on imbalanced CIFAR datasets and two large-scale datasets with a long-tailed distribution.
From the experimental results, we confirmed that the proposed loss function was much improved, in comparison with LDAM loss and conventional loss functions for class-imbalanced learning.
Additionally, by comparing the feature space between conventional margin-based losses and our proposed loss, we confirmed ELM loss can create a better feature space where the rare classes are more distant from each frequency class. 
Our code is available at \footnote{\url{https://github.com/usagisukisuki/ELMloss}}.

The main contributions of this paper are as follows:
 \begin{itemize}
  \item  We reconsider the original LDAM loss in terms of a large-margin perspective.
  As a result, we confirm that LDAM loss includes wider and class-specific values of large margins than LMSCE loss, and we consider LDAM loss is a good performance under imbalanced learning.
  \item  Furthermore, we propose a novel loss function with even wider margins than LDAM loss, called Enlarged Large Margin (ELM) loss.
  ELM loss has a stronger regularization than LDAM loss, and its regularization can consist of further increasing the margins between the logit of the target class and the maximum logit of the other class.
\end{itemize}

This paper is organized as follows. 
Section $\rm{I\hspace{-1.2pt}I}$ describes the related works. 
Section $\rm{I\hspace{-1.2pt}I\hspace{-1.2pt}I}$ describes our methodology.
Section $\rm{I\hspace{-1.2pt}V}$ presents the experimental results. 
Finally, we describe our summary and future works in Section $\rm{V}$.


\section{RELATED STUDIES}
\begin{figure}[t]
\begin{center}
\includegraphics[scale=0.5]{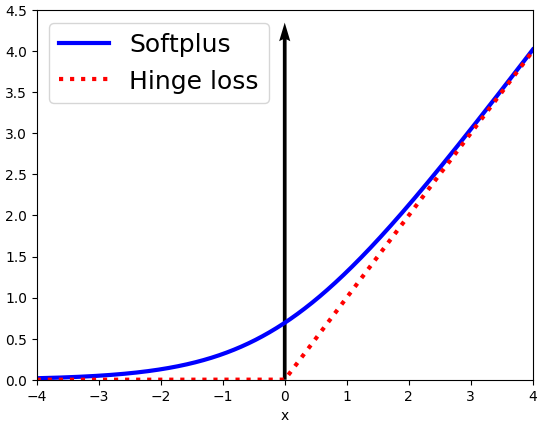}
\end{center}
\caption{Comparison of hinge loss and softplus loss.}
\end{figure}
\subsection{Margin based loss}
A margin-based approach is typically obtained by using the hinge loss function such as SVM \cite{b16}, for improving the generalization performance, while mitigating the over-fitting. 
The classifier benefits from margins to maximize the inter-class margin.
Especially, following \cite{b17}, it is general to regard as a classification margin the difference between the logit for the target class and the maximum logit for the other class.
In the field of deep learning, unique loss functions with the benefits of the margin have been proposed \cite{b15,b18,b19,b20}. 
In particular, it has been revealed that the original softmax cross entropy loss includes a large-margin effect that is dependent on the distribution of logits \cite{b15}, and it is formulated the regularization term to further enhance the large-margin effect in softmax cross entropy loss.

In contrast to the above loss functions with class-independent margins, another approach, which uses different margins for each class, has been proposed \cite{b4,b6}.
Cao \textit{et al}.\cite{b4} proposed Label-Distribution-Aware Margin (LDAM) loss that minimizes a margin-based generalization bound.
The margin is obtained from the number of data divided by the $\frac{1}{4}$ power of the number of samples for each class.
By using LDAM loss, it is possible to obtain large margins for the categories with a small number of samples and small margins for the categories with a large number of samples.
Haeyong \textit{et al}.\cite{b6} proposed a novel Maximum Margin (MM) loss function to encourage larger sample margins for hard negative sample classes such that the smaller the maximum margins are the greater the shifting margins are.

However, even though these margin losses for imbalanced classification are defined on the basis of softmax cross entropy loss, the relation to the large margin included in softmax cross entropy loss has not been discussed.
In this study, we reconsider the connection between LDAM loss and LMSCE loss using the softplus function and confirm that LDAM loss is assigned wider margins than LMSCE loss.

\subsection{Re-weighting approach}
The re-weighting approach is learned by imposing a large weight on unacceptable errors and a small weight on acceptable errors \cite{b3,b4,b5,b6,b7,b8,b9,b10,b11,b12,b21}. 
Class-balancing weight \cite{b3} is one of the common approaches for re-weighting. 
This method calculates the weights based on the number of training samples in each class and weights the loss function to increase the penalty for minority classes.
By using class-balancing weight, it is possible to balance the learning.
Furthermore, the Deferred Re-balancing Weighting (DRW) strategy, which trains using the vanilla loss before annealing the learning rate and deploys a class-balancing weight with a smaller learning rate, has also been proposed \cite{b4}.

As another approach, novel loss functions to consider the distribution of class labels have been proposed \cite{b4,b5,b6,b7,b8,b9,b10,b11,b12}.
Focal loss \cite{b12}, Equalization losses \cite{b7,b8}, Group softmax \cite{b9}, and Adaptive Class Suppression loss \cite{b10} are typical methods.

Although various loss functions for imbalanced classification have been proposed, the loss using both LDAM loss and DRW is known to be effective on many imbalanced datasets and a lot of tasks other than image classification.
Our proposed loss function is a development of LDAM loss which provides higher performance when we used the combination with the DRW strategy.

\section{METHODOLOGY}
\begin{figure*}[t]
    \begin{tabular}{ccc}
      \begin{minipage}{0.345\hsize}
        \centering
        \includegraphics[scale=0.5]{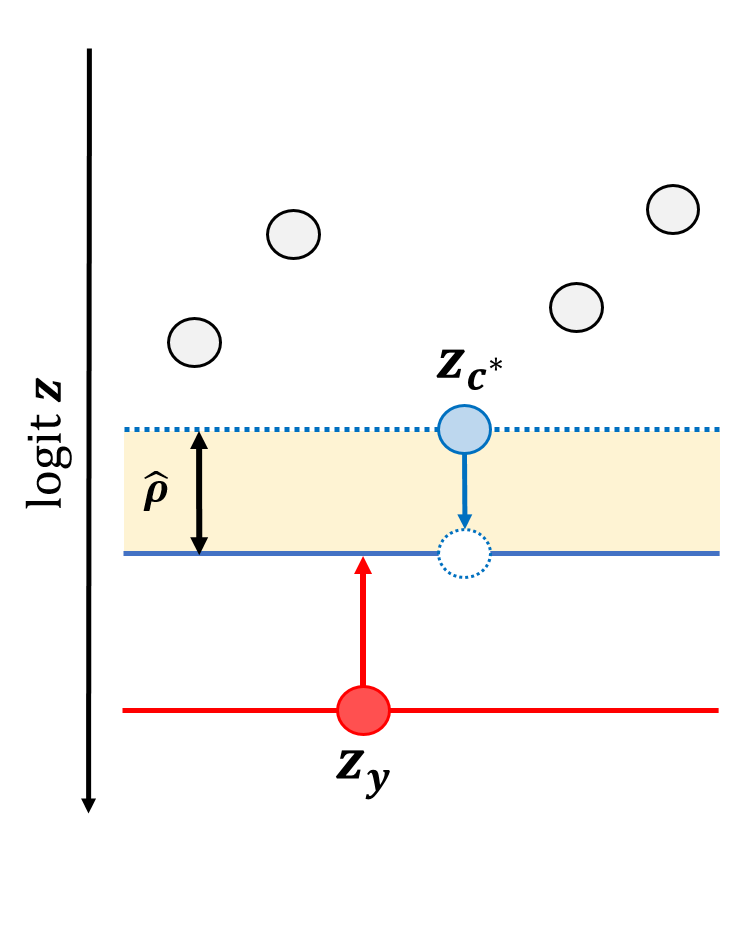}
        \subcaption{LMSCE}
    \end{minipage}%
    \begin{minipage}{0.35\hsize}
        \centering
        \includegraphics[scale=0.5]{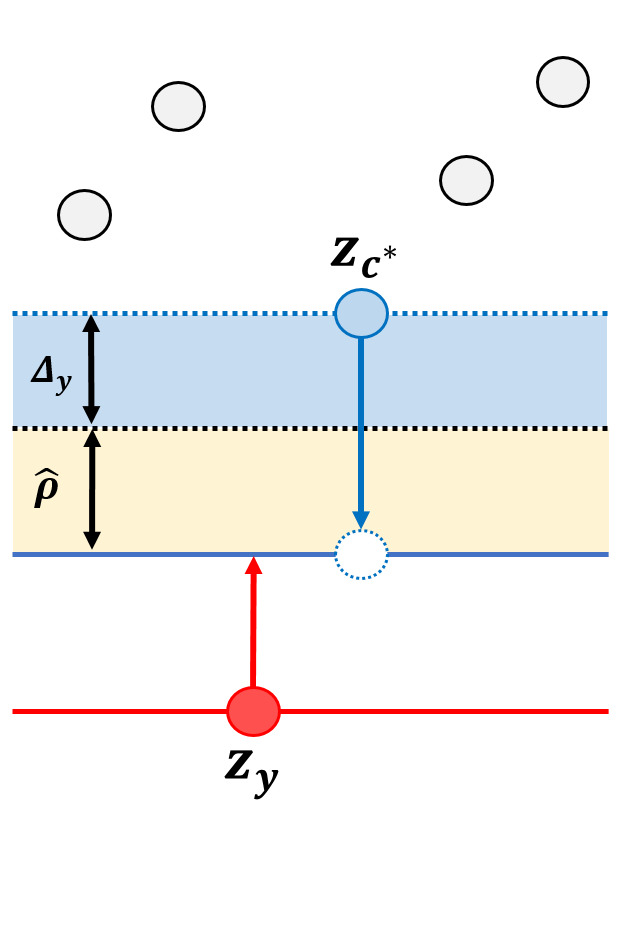}
        \subcaption{LDAM}
    \end{minipage}%
    \begin{minipage}{0.29\hsize}
        \centering
        \includegraphics[scale=0.5]{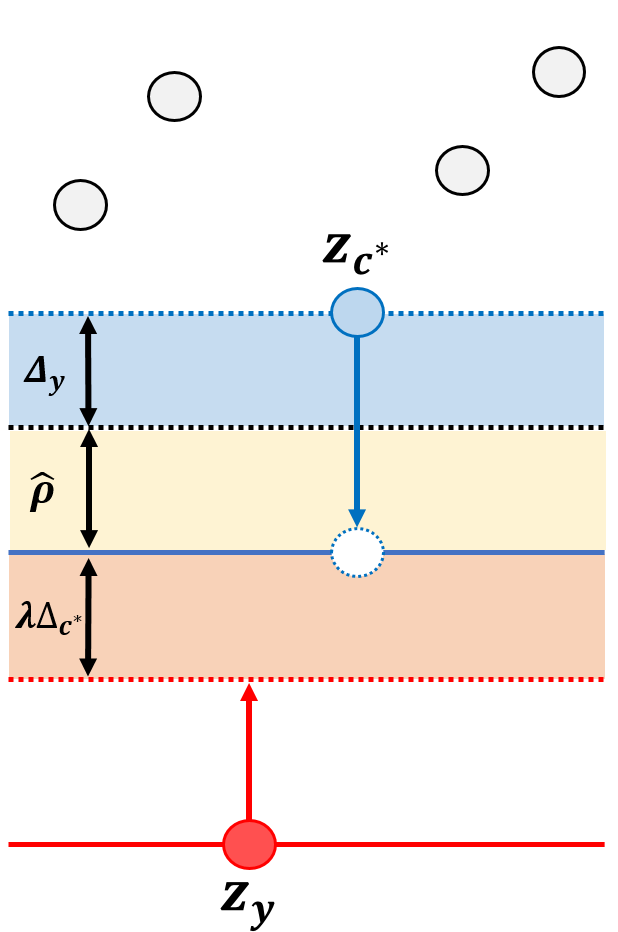}
        \subcaption{ELM (Ours)}
    \end{minipage}%
    \end{tabular}
    \caption{Comparison between conventional margin-based loss functions and our proposed loss function.}
\end{figure*}
In this section, we reconsider LDAM loss in terms of a large-margin perspective and propose a novel strategy to utilize a wider large margin, called ELM loss.
By using this approach, the model can learn with stronger regularization than LDAM loss, and we can achieve better-balanced learning than LDAM loss even if the dataset has a class-imbalanced distribution.

\subsection{Large-margin for LDAM loss}
In the multi-class classification of $C$ classes, a CNN model provides a logit vector $\textbf{\textit{z}}\in\mathbb{R}^C$. 
The logit vector is generally produced by the last fully-connected layer applying a linear classifier to the feature vector $\textbf{\textit{x}}$ at the penultimate layer. 
Given the logit vector and the ground truth label $y\in \lbrace1,..., C\rbrace$, the original softmax cross entropy loss is formulated as
\begin{eqnarray}
  Loss(\textbf{\textit{z}},y)= - \log \frac{e^{z_y}}{\sum_{c=1}^C {e^{z_c}}} = \log\biggl\lbrack1+\sum_{c\neq y}^C e^{(z_c - z_y)}\biggl\rbrack
\end{eqnarray}
Kobayashi \cite{b15} indicated that Eq. (1) can be converted to Eq. (2) by applying the softplus function, which is described by $softplus(x)=log(1+e^x)$.
\begin{gather}
  Loss(\textbf{\textit{z}},y) = softplus\biggl\lbrack \log\Bigl\lbrace \sum_{c\neq y} e^{z_c} \Bigl\rbrace - z_y	\biggl\rbrack
\end{gather}
Subsequently, we apply to the log-sum-exp transformation that aggregates the logits of the nontarget classes $c\neq y$ in Eq. (2), and the result of the conversion indicates in Eq. (3).
\begin{gather}
  Loss(\textbf{\textit{z}},y) = softplus(z_{c^*}-z_y+\bm{\hat{\rho}})\\
  \bm{\hat{\rho}} = \log\biggl\lbrack\sum_{c\neq y} e^{(z_c-z_{c^*})}\biggl\rbrack\notag
\end{gather}
where $z_{c^*}$ is the maximum logit of incorrect class and $\hat{\rho}$ is a bias margin.
The softplus function, as shown in Fig. 1, is similar to the hinge loss. 
This formulation reveals that the softmax cross entropy loss measures the significance of the target logit $z_y$ compared with the other logit $z_{c^*}$.
Then, the original softmax cross entropy loss induces a large-margin classifier depending on the bias $\hat{\rho}$, and we call this LMSCE loss.

Inspired by this conversion method using the softplus function, we convert LDAM loss described in Eq. (4) to the large-margin loss in Eq. (5).
\begin{eqnarray}
  Loss_{ldam}(\textbf{\textit{z}},y) = - \log \frac{e^{s(z_y-\Delta_{y})}}{e^{s(z_y-\Delta_{y})} + \sum_{j\neq y} {e^{z_j}}}
\end{eqnarray}
\begin{gather}
  Loss_{ldam}(\textbf{\textit{z}},y) = softplus(s(z_{c^*}-z_y) + s\Delta_{y}+\bm{\hat{\rho}})\\
  \bm{\hat{\rho}} = \log\biggl\lbrack\sum_{c\neq y} e^{s(z_c-z_{c^*})}\biggl\rbrack\notag
\end{gather}
where $\Delta_{y}$ is a class-dependent margin, which is described by $\Delta_{j}=\frac{M}{n_{j}^{1/4}}\:for\:j\in\lbrace1,..,k..,C\rbrace$.
$n_{j}$ is the number of training samples in class $j$, $M$ is the maximum size of margin that is tuned as a hyper-parameter, and $s$ is a scaling factor.
As shown in Eq. (5), LDAM loss can be converted by the large margin framework as well as LMSCE loss.
Additionally, compared between LDAM loss and LMSCE loss in Fig. 2 (a-b), LDAM loss is given a wider margin $\Delta_{j}$ than LMSCE loss, and the range of margins vary from class.

\subsection{Enlarged Large Margin loss}
\begin{table*}[t]
    \centering
    \caption{Comparison of the results on imbalanced CIFAR datasets.}
    \scalebox{1.1}{
    \begin{tabular*}{16cm}{@{\extracolsep{\fill}}rcccccccc} \bhline{1.0pt}
    \multicolumn{1}{r}{} & \multicolumn{2}{c}{CIFAR10-LT} &  \multicolumn{2}{c}{CIFAR10-Step}&\multicolumn{2}{c}{CIFAR100-LT} & \multicolumn{2}{c}{CIFAR100-Step}\\
    \cmidrule(lr){2-3}%
    \cmidrule(lr){4-5}%
    \cmidrule(lr){6-7}%
    \cmidrule(lr){8-9}%
    \multicolumn{1}{r}{Loss function} & Ratio=10 & Ratio=100 & Ratio=10 & Ratio=100 & Ratio=10 & Ratio=100 & Ratio=10 & Ratio=100\\
    \hline
        CE &86.92\tiny{±0.06}&70.71\tiny{±0.20}&84.44\tiny{±0.35}&64.36\tiny{±0.30}&56.54\tiny{±0.35}&38.54\tiny{±0.41}&54.48\tiny{±0.12}&38.71\tiny{±0.19}\\
        LMSCE &87.03\tiny{±0.24}&72.34\tiny{±0.77}&85.08\tiny{±0.23}&62.29\tiny{±0.63}&56.44\tiny{±0.16}&36.90\tiny{±0.14}&50.70\tiny{±0.50}&38.89\tiny{±0.11}\\
        Focal &86.41\tiny{±0.18}&71.59\tiny{±0.53}&84.54\tiny{±0.28}&66.13\tiny{±2.46}&55.98\tiny{±0.19}&38.39\tiny{±0.52}&54.88\tiny{±0.36}&38.76\tiny{±0.18}\\
        Equalization v2 &86.14\tiny{±0.28}&69.49\tiny{±0.12}&84.11\tiny{±0.59}&61.40\tiny{±1.58}&56.37\tiny{±0.10}&38.09\tiny{±0.52}&53.42\tiny{±0.32}&38.35\tiny{±0.04}\\
        IB FLocal &82.62\tiny{±0.52}&68.82\tiny{±0.52}&87.87\tiny{±0.19}&74.78\tiny{±0.63}&56.52\tiny{±0.32}&40.61\tiny{±0.71}&56.94\tiny{±0.36}&43.70\tiny{±0.49}\\
        LDAM &86.68\tiny{±0.25}&73.67\tiny{±0.29}&84.55\tiny{±0.39}&65.21\tiny{±1.05}&55.38\tiny{±0.25}&39.82\tiny{±0.45}&50.79\tiny{±0.37}&39.37\tiny{±0.13}\\
        MM-LDAM &86.23\tiny{±0.24}&71.51\tiny{±0.48}&83.86\tiny{±0.25}&56.20\tiny{±1.18}&53.25\tiny{±0.30}&36.89\tiny{±0.25}&46.34\tiny{±0.32}&38.13\tiny{±0.06}\\
        LDAM + DRW &87.36\tiny{±0.28}&77.16\tiny{±0.33}&87.65\tiny{±0.27}&77.59\tiny{±0.14}&57.71\tiny{±0.32}&42.97\tiny{±0.62}&57.03\tiny{±0.09}&45.75\tiny{±0.16}\\
        MM-LDAM + DRW&86.63\tiny{±0.17}&77.79\tiny{±0.21}&86.87\tiny{±0.08}&76.31\tiny{±0.43}&51.13\tiny{±0.42}&39.31\tiny{±0.38}&48.35\tiny{±0.27}&40.74\tiny{±0.42}\\
    \hline
        ELM w/o $\Delta_{y}$ ($\lambda$=1.0)&47.30\tiny{±2.01}&60.37\tiny{±0.67}&53.89\tiny{±6.27}&64.14\tiny{±0.20}&17.69\tiny{±3.68}&33.63\tiny{±1.21}&36.64\tiny{±4.96}&37.98\tiny{±0.41}\\
        ELM w/o $\Delta_{y}$ ($\lambda$=0.5)&84.12\tiny{±0.18}&72.44\tiny{±0.43}&83.68\tiny{±0.67}&68.06\tiny{±0.83}&51.97\tiny{±0.49}&37.10\tiny{±0.81}&51.32\tiny{±0.84}&39.32\tiny{±0.21}\\
        ELM w/o $\Delta_{y}$ ($\lambda$=0.1)&86.76\tiny{±0.08}&71.61\tiny{±1.08}&85.06\tiny{±0.03}&65.61\tiny{±1.08}&55.64\tiny{±0.38}&38.04\tiny{±0.39}&53.69\tiny{±0.65}&39.47\tiny{±0.23}\\
    \hline
        ELM ($\lambda$=1.2)&83.40\tiny{±0.49}&76.35\tiny{±0.63}&75.77\tiny{±0.48}&74.73\tiny{±0.46}&53.88\tiny{±0.83}&40.14\tiny{±0.71}&54.49\tiny{±0.59}&43.32\tiny{±0.27}
\\
        ELM ($\lambda$=1.0)&87.61\tiny{±0.04}&77.61\tiny{±0.86}&87.53\tiny{±0.35}&76.09\tiny{±0.90}&57.45\tiny{±0.45}&40.93\tiny{±0.82}&57.59\tiny{±0.46}&43.37\tiny{±0.83}\\
        ELM ($\lambda$=0.5)&87.82\tiny{±0.11}&75.14\tiny{±0.47}&87.38\tiny{±0.17}&71.26\tiny{±0.85}&58.30\tiny{±0.19}&41.61\tiny{±0.86}&57.01\tiny{±0.32}&40.82\tiny{±0.31}\\
        ELM ($\lambda$=0.1)&87.05\tiny{±0.07}&73.34\tiny{±0.84}&84.93\tiny{±0.20}&66.44\tiny{±0.46}&56.59\tiny{±0.06}&40.36\tiny{±0.45}&51.93\tiny{±0.23}&39.62\tiny{±0.12}\\
        ELM + DRW ($\lambda$=1.2)&81.82\tiny{±0.27}&77.57\tiny{±0.26}&72.62\tiny{±0.40}&77.06\tiny{±1.14}&51.92\tiny{±0.72}&40.07\tiny{±1.32}&52.15\tiny{±0.84}&43.45\tiny{±0.84}\\
        ELM + DRW ($\lambda$=1.0)&88.16\tiny{±0.17}&\textbf{78.89\tiny{±0.03}}&88.42\tiny{±0.11}&78.03\tiny{±0.16}&\textbf{58.93\tiny{±0.33}}&41.96\tiny{±0.33}&58.17\tiny{±0.60}&43.75\tiny{±0.24}\\
        ELM + DRW ($\lambda$=0.5)&\textbf{88.31\tiny{±0.08}}&78.05\tiny{±0.13}&\textbf{88.67\tiny{±0.17}}&\textbf{78.17\tiny{±0.29}}&58.82\tiny{±0.27}&42.35\tiny{±1.16}&\textbf{60.13\tiny{±0.12}}&45.74\tiny{±0.33}\\
        ELM + DRW ($\lambda$=0.1)&87.68\tiny{±0.32}&77.84\tiny{±0.61}&88.03\tiny{±0.09}&78.17\tiny{±0.49}&58.35\tiny{±0.15}&\textbf{43.07\tiny{±0.55}}&57.74\tiny{±0.60}&\textbf{46.00\tiny{±0.18}}\\
    \bhline{1.0pt} 
    \end{tabular*}
    }
\end{table*}
In Section $\rm{I\hspace{-1.2pt}I\hspace{-1.2pt}I}$-A, we confirmed that LDAM loss is given wider and class-specific larger margins than LMSCE loss. 
We consider that LDAM loss can be balanced learning for the imbalanced dataset because it includes a large margin depending on the label distribution. 
However, LDAM loss utilizes only the large margin for the correct class, not the large margin for the maximum logit of the incorrect class.
As shown in Eq. (5), since LDAM loss implicitly learns the difference between $z_{c^*}$ and $z_y$, we consider that LDAM loss should also be used in the large margin for the maximum logit $z_{c^*}$.
Therefore, we propose a novel Enlarged Large Margin (ELM) loss to adopt the large margin toward $z_{c^*}$.
ELM loss defined by the softplus function is shown in Eq. (6).
\begin{gather}
  Loss_{elm}(\textbf{\textit{z}},y) = softplus(s(z_{c^*}-z_y) + s\Delta_{y} - s\lambda\Delta_{c^*}+\bm{\hat{\rho}})\\
  \bm{\hat{\rho}} = \log\biggl\lbrack\sum_{c\neq y} e^{s(z_c-z_{c^*})}\biggl\rbrack\notag
\end{gather}
where $\Delta_{c^*}$ is the margin for the maximum logit of the other class, and it is defined by the same calculation as the margin of LDAM loss.
$\lambda$ is a hyper parameter that determine the strength of $\Delta_{c^*}$.
When we return Eq. (6) defined by the softplus function to the usual cross-entropy loss form, the final loss function is indicated in Eq. (7). 
\begin{eqnarray}
  Loss_{elm}(\textbf{\textit{z}},y) = - \log \frac{e^{s(z_y-\Delta_{y}+\lambda\Delta_{c^*})}}{e^{s(z_y-\Delta_{y}+\lambda\Delta_{c^*})} + \sum_{j\neq y} {e^{z_j}}}
\end{eqnarray}

As shown in Fig. 2 (c), ELM loss can take a wider large margin for $\lambda\Delta_{c^*}$ compared to the conventional LMSCE loss and LDAM loss.
When the correct class is the rare class in imbalanced classification, $z_{c^*}$ often takes the logit of the frequent class.
Consequently, the large margin $\Delta_{y}$ used in LDAM loss can be taken, allowing for greater distance separation.
In contrast, in ELM loss, the large margin $\lambda\Delta_{c^*}$ can be used in addition to $\Delta_{y}$, and the distance between $z_{c^*}$ and $z_y$ can be greater than LDAM loss.
Further, in the imbalance dataset, the most frequently occurring sample in the mini-batch is the majority class, and then LDAM loss has employed the small and the same value of $\Delta_{y}$ in most cases.
In contrast, since EML loss can use the margins of classes $z_{c^*}$, more variety of margins can be used for majority classes.

\section{EXPERIMENTS}
\begin{table}[t]
    \centering
    \caption{Comparison of the results on ImageNet-LT dataset.}
    \scalebox{1.15}{
    \begin{tabular*}{7.0cm}{@{\extracolsep{\fill}}rccc} \bhline{1.0pt}
    \multicolumn{1}{r}{} & \multicolumn{3}{c}{Accuracy}\\
    \cmidrule(lr){2-4}%
    \multicolumn{1}{r}{Loss function} & Top 1 & Top 3 & Top 5\\
    \hline 
        CE &47.66&66.87&74.19 \\
        LMSCE &45.57&64.55&71.61 \\
        Focal &47.76&66.68&73.68 \\
        Equalization v2 &47.84&67.22&74.60 \\
        IB Focal &47.77&67.65&74.88\\
        LDAM &50.53&69.20&75.64\\
        MM-LDAM &43.33&63.02&70.46\\
        LDAM + DRW &53.38&69.36&75.35 \\
        MM-LDAM + DRW&47.74&65.04&71.38\\
    \hline
        ELM ($\lambda$=1.0)&\textbf{55.49}&\textbf{71.88}&\textbf{77.58}\\
        ELM ($\lambda$=0.5)&54.24&71.42&77.22\\
        ELM ($\lambda$=0.1)&51.50&69.53&76.11\\
        ELM + DRW ($\lambda$=1.0)&51.22&71.10&77.23\\
        ELM + DRW ($\lambda$=0.5)&54.50&70.98&76.40\\
        ELM + DRW ($\lambda$=0.1)&53.68&69.88&75.72\\
    \bhline{1.0pt} 
    \end{tabular*}
    }
\end{table}
\begin{table}[t]
    \centering
    \caption{Comparison of the results on Places-LT dataset.}
    \scalebox{1.15}{
    \begin{tabular*}{7.0cm}{@{\extracolsep{\fill}}rccc} \bhline{1.0pt}
    \multicolumn{1}{r}{} & \multicolumn{3}{c}{Accuracy}\\
    \cmidrule(lr){2-4}%
    \multicolumn{1}{r}{Loss function} & Top 1 & Top 3 & Top 5\\
    \hline
        CE &26.63&47.91&58.48 \\
        LMSCE &24.34&44.70&54.83 \\
        Focal &26.88&48.07&58.21 \\
        Equalization v2 &26.75&49.13&59.90 \\
        IB FLocal &32.51&\textbf{54.86}&\textbf{65.12} \\    
        LDAM &26.28&46.47&56.06\\
        MM-LDAM &22.87&44.51&54.94\\
        LDAM + DRW &31.81&50.81&59.49 \\
        MM-LDAM + DRW&30.58&50.97&60.39\\
    \hline
        ELM ($\lambda$=1.0)&31.47&51.15&60.28\\
        ELM ($\lambda$=0.5)&28.42&48.57&58.46\\
        ELM ($\lambda$=0.1)&26.66&47.15&57.51\\
        ELM + DRW ($\lambda$=1.0)&32.16&53.49&63.07\\
        ELM + DRW ($\lambda$=0.5)&\textbf{33.02}&53.15&62.17\\
        ELM + DRW ($\lambda$=0.1)&32.61&52.06&61.05\\
    \bhline{1.0pt} 
    \end{tabular*}
    }
\end{table}

\subsection{Datasets and learning conditions}
\noindent
\textbf{CIFAR}. The original version of CIFAR-10 and CIFAR-100 contains 5,000 and 500 training images of $32 \times 32$ pixels with 10 and 100 classes, respectively.
Following \cite{b3}, we reduced the number of training images and make long-tailed and step distributions.
We used the imbalanced ratio to denote between the samples of the most frequent and the least frequent categories and make two types of datasets with different imbalance ratios $(e.g., 10, 100)$.
Evaluation images comprise original test images.
For data pre-processing, training samples were randomly cropped to $32 \times 32$ in the case that padding size is 4, and flipped horizontally.

We aligned all training conditions with the rule of \cite{b3}.
The network was the ResNet32 \cite{b3} with full-scratch learning. 
The batch size is set to 128, the epochs are set to 200, and the optimizer is Stochastic Gradient Descent (SGD) with a momentum of 0.9 and weight decay of $2\times10^{-4}$.
The initial learning rate was 0.1, and we use a standard learning schedule that was decayed by 0.01 at the 160 epoch and again at the 180 epoch.
We also used a warm-up learning rate schedule for the first 5 epochs.
Experiments were conducted three times by altering the random initial values of the parameters.
The average accuracy of the three times experiments was used for the evaluation. 

\begin{figure*}[t]
    \centering
    \begin{tabular}{ccc}
      \begin{minipage}{0.31\hsize}
          \centering
        \includegraphics[scale=0.26]{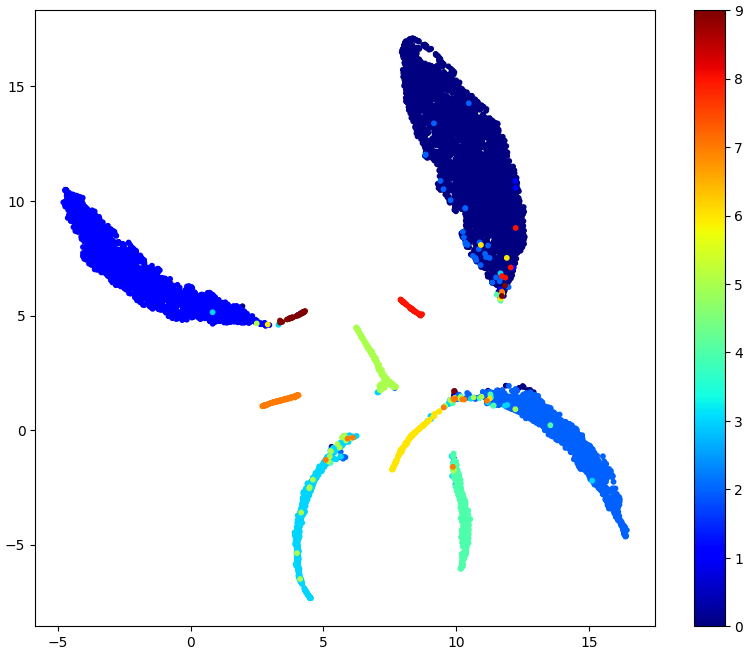}
        \subcaption{LDAM }
    \end{minipage}%
    \begin{minipage}{0.31\hsize}
        \centering
        \includegraphics[scale=0.26]{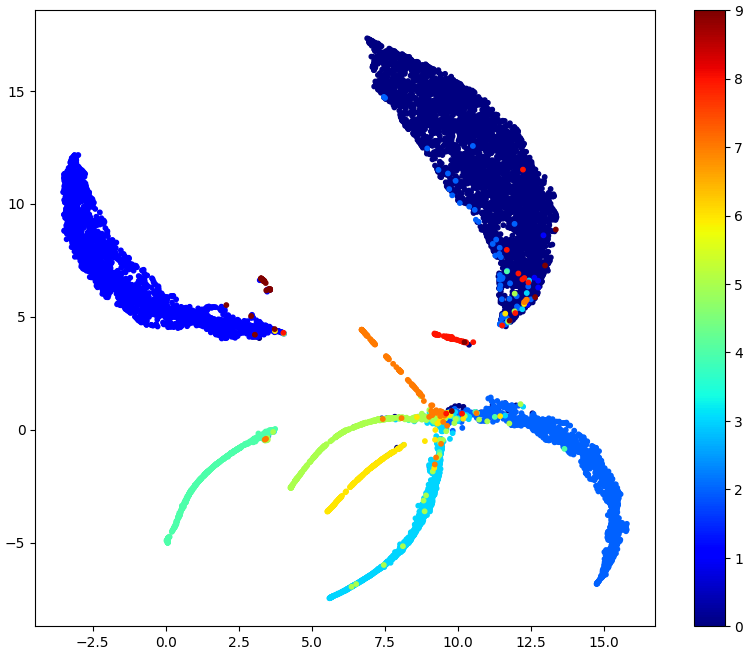}
        \subcaption{MM-LDAM}
    \end{minipage}%
    \begin{minipage}{0.31\hsize}
        \centering
        \includegraphics[scale=0.26]{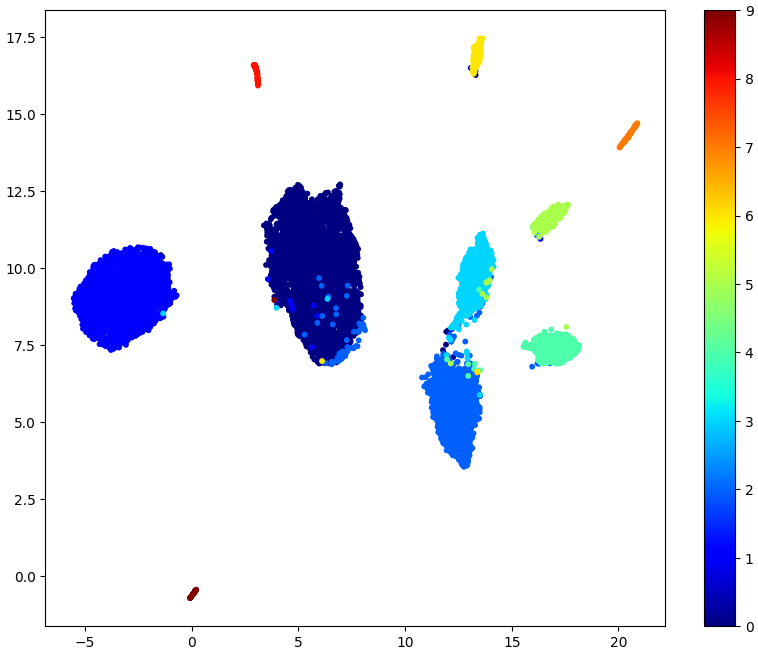}
        \subcaption{ELM (Ours)}
    \end{minipage}%
    \end{tabular}
    \caption{Visualization of penultimate layer in ResNet32 trained on the CIFAR10-LT with ratio=100. The color bar shows the class number and indicates a small number of samples as the class number increase.}
\end{figure*}

\noindent
\textbf{Large-scale dataset}. We used two types of common large-scale imbalanced datasets, ImageNet-LT \cite{b1}, and Places-LT \cite{b1}.
ImageNet-LT \cite{b1} contains 115,800 images for 1,000 categories, with a maximum of 1,280 images per category and a minimum of 5 images per category.
It is sampled from ImageNet-2012\cite{b13} following the Pareto distribution.
Further, a long-tailed version of Places-2 \cite{b22} is constructed in a similar way. 
Places-LT \cite{b1} contains 184.5K images for 365 categories, with a maximum of 4,980 images per category and a minimum of 5 images per category.
The gap between the head and tail classes is even larger than ImageNet-LT.

For data pre-processing, training samples were randomly cropped to $224\times224$, flipped horizontally, and randomly changed brightness values.
In inference, samples were resized to $256\times256$ and cropped $224\times224$ in the center.
We used the ResNet50 pre-trained on the ImageNet \cite{b13} as following \cite{b4}. 
We set the batch size to 256 and the number of epochs to 90. 
The optimizer was the SGD with momentum and weight decay ({\sl $lr$}=$1\times10^{-1}$, $momentum=0.9$, $2\times10^{-4}$).
We used a standard learning rate schedule that is decayed by 0.1 at 50, 70, and 85 epochs, and a warm-up learning rate schedule for the first 5 epochs for training.

Experiments were conducted one time by altering the random initial values of the parameters, and we used accuracy, which is the percentage of images whose labels are at the top predicted by the network (top-1, top-3, and top-5 accuracy) as the evaluation metric.


\subsection{Results}
As comparison methods, we employed the standard softmax cross entropy (CE) loss and conventional losses for imbalanced classification, which are Focal loss \cite{b12}, Equalization v2 loss \cite{b8}, and IB Focal loss \cite{b21}.
Furthermore, we also compared conventional margin-based losses, which are LMSCE loss \cite{b18}, LDAM loss \cite{b4}, and MM-LDAM loss \cite{b6}.
Table 1 presents the evaluation results on test images in the imbalanced CIFAR datasets\cite{b3}.
The bold letters show the best accuracy.
In imbalanced CIFARs, ELM loss was better performance compared with conventional methods and ELM loss with DRW was the highest accuracy in all ratios.
Even though ELM loss was a simple expansion method of LDAM loss, it was effective in comparison with the conventional loss functions for imbalanced classification.
Additionally, by setting a smaller $\lambda$, the proposed large margin could be used more effectively because ELM loss extends the margins further in addition to the original LDAM margin, and the margins may become too large depending on the number of training samples in the dataset.
Moreover, when only the margins of ELM loss were used without the margins originally used for LDAM loss (ELM w/o $\Delta_y$, in Table 1), the accuracy was also better than that of CE loss and LMSCE loss by setting small $\lambda$.

Table 2 and Table 3 present the evaluation results on test images in the ImageNet-LT and Places-LT datasets \cite{b1}.
In ImageNet-LT, EML loss had a higher accuracy than other margin-based losses, and improved top1 accuracy by over $4.9\%$ than LDAM loss.
Furthermore, EML loss with DRW was the highest accuracy in comparison with conventional approaches for the imbalanced dataset.
In Places-LT, ELM loss was also the better performance compared to conventional margin-based methods with top1 accuracy.
These results demonstrate that ELM loss is an effective generalizable method on a variety of imbalanced datasets.
In ImageNet-LT, the maximum number of images is 1,280 and the minimum number of images is 5, and the imbalance ratio is larger than in imbalanced CIFAR and there are classes where the margins are too small. 
We consider that the margins with $\lambda=1.0$ work the best in that case.
Additionally, it is possible that a class balancing weight is too strong for the large-scale dataset if EML loss with DRW is used, and generalization performance may decrease due to over-fitting. 
In that case, we consider using only ELM loss is the best performance.

Figure 3 illustrates the visualization results of features at a penultimate layer that is compressed to two dimensions by UMAP\cite{b23} when we evaluated training samples. 
The color bar shows the class number and indicates a small number of samples as the class number increase. 
As shown in Fig. 3, in conventional loss functions such as LDAM loss and MM-LDAM loss, the distance between minority classes and other classes is quite close, and the feature spaces of minority classes are not independent. 
However, when we employed ELM loss, the feature space is more effective for discrimination since the distance between the minority class and the other classes is more independent.
These results indicate that ELM loss has a stronger regularization than LDAM loss and other margin-based losses and demonstrate the effectiveness of increasing the margins between the logit of the correct class and the maximum logit of the wrong class.

\section{CONCLUSION}
In this paper, we reconverted a formula for LDAM loss in terms of a large-margin perspective and proposed a novel ELM loss function with even wider margins than LDAM loss.
Through experiments on imbalanced classification, ELM loss significantly improved the accuracy compared with conventional large-margin losses.
Furthermore, we confirmed that ELM loss generated effective feature space for classification.

To further improve the accuracy
of minority classes, while maintaining the accuracy of majority classes, the combination of EML loss with effective strategies for long-tailed classification such as two-stage learning is one of our future works.


\end{document}